\documentclass[letterpaper, 10 pt, conference]{ieeeconf}  

\IEEEoverridecommandlockouts                              

\usepackage[table,xcdraw]{xcolor}
\usepackage{etoolbox} 
\usepackage[utf8]{inputenc} 
\usepackage[T1]{fontenc}    
\usepackage[english]{babel} 
\usepackage[protrusion=true,expansion=true]{microtype}
\usepackage{comment}
\usepackage{cancel}
\usepackage{csquotes}
\usepackage{listings}
\usepackage{nameref}
\usepackage{paralist}
\usepackage{xspace}
\usepackage{makeidx}
\usepackage{pdfpages}
\usepackage{footnote}

\usepackage{hyperref}
\hypersetup{
    colorlinks=true,
    linkcolor=black,
    citecolor=black,
    filecolor=cyan,
    urlcolor=blue
}

\usepackage{flushend}
\usepackage{balance}

\usepackage{paralist}
\let\labelindent\relax 
\usepackage{enumitem}
\setlist[itemize]{noitemsep,leftmargin=*,topsep=0in}
\setlist[enumerate]{noitemsep,leftmargin=*,topsep=0in}

\usepackage{amsmath,amssymb} 
\usepackage{amsfonts}       
\usepackage{mathtools}
\usepackage{array} 
\usepackage{nicefrac}       
\usepackage{breqn}          
\usepackage{textcomp}
\usepackage{bm} 
\usepackage{pifont}
\usepackage[
  separate-uncertainty = true,
  multi-part-units = repeat
]{siunitx}

\usepackage{graphicx}
\usepackage{adjustbox} 
\usepackage{epsfig}

\usepackage{float}
\usepackage[font=small]{subcaption}
\usepackage[font=small,labelfont=bf]{caption}
\usepackage{lscape}      
\usepackage{wrapfig}

\usepackage{tikz}  
\usepackage{makecell}
\usepackage{overpic}
\usepackage{algorithm}
\usepackage[noend]{algpseudocode}

\usepackage{tabularx}
\usepackage{multirow}
\usepackage{multicol}
\usepackage{booktabs}   
\usepackage{array} 

\usepackage{longtable}
\usepackage{threeparttable}

\makesavenoteenv{tabular}
\makesavenoteenv{table}

\usepackage{cite}

\renewcommand{\paragraph}[1]{\vspace{0.2em}\noindent\textit{#1} --}

\newcommand\mybar{\kern1pt\rule[-\dp\strutbox]{.8pt}{\baselineskip}\kern1pt}

\newcommand{\nvidia}{\textsc{Nvidia}\xspace}
\newcommand{\simName}{\textsc{Orbit}-Surgical\xspace}
\newcommand{\this}{\textsc{SuFIA-BC}\xspace}

\title{\LARGE \bf
\this: Generating High Quality Demonstration Data for Visuomotor Policy Learning in Surgical Subtasks
}

\author{
Masoud Moghani$^{1,2}$,
Nigel Nelson$^{2}$,
Mohamed Ghanem$^{3}$,
Andres Diaz-Pinto$^{2}$,
Kush Hari$^{4}$, \\
Mahdi Azizian$^{2}$,
Ken Goldberg$^{4}$,
Sean Huver$^{2}$,
Animesh Garg$^{1,2,3}$
\thanks{$^{1}$University of Toronto, $^{2}$NVIDIA, $^{3}$Georgia Institute of Technology, $^{4}$University of California, Berkeley}%
\thanks{Correspondence to: \href{mailto:moghani@cs.toronto.edu}{moghani@cs.toronto.edu}, \href{mailto:garg@cs.toronto.edu}{garg@cs.toronto.edu}}%
}

\begin{document}

\maketitle
\thispagestyle{empty}
\pagestyle{empty}

\begin{abstract}

Behavior cloning facilitates the learning of dexterous manipulation skills, yet the complexity of surgical environments, the difficulty and expense of obtaining patient data, and robot calibration errors present unique challenges for surgical robot learning. We provide an enhanced surgical digital twin with photorealistic human anatomical organs, integrated into a comprehensive simulator designed to generate high-quality synthetic data to solve fundamental tasks in surgical autonomy. We present \this: visual Behavior Cloning policies for Surgical First Interactive Autonomy Assistants. We investigate visual observation spaces including multi-view cameras and 3D visual representations extracted from a single endoscopic camera view. Through systematic evaluation, we find that the diverse set of photorealistic surgical tasks introduced in this work enables a comprehensive evaluation of prospective behavior cloning models for the unique challenges posed by surgical environments. We observe that current state-of-the-art behavior cloning techniques struggle to solve the contact-rich and complex tasks evaluated in this work, regardless of their underlying perception or control architectures. These findings highlight the importance of customizing perception pipelines and control architectures, as well as curating larger-scale synthetic datasets that meet the specific demands of surgical tasks.

\noindent Project website: \href{https://orbit-surgical.github.io/sufia-bc/}{orbit-surgical.github.io/sufia-bc/}

\end{abstract}

\section{Introduction}

The adoption of robotic surgical assistants (RSAs) in operating rooms offers significant benefits to both surgeons and patient outcomes. Current surgical robot platforms are controlled via teleoperation through a console by a trained surgeon. Augmented dexterity in surgery has the potential to simplify the surgical workflow while relieving surgeons’ workload \cite{kyg}.

Surgical autonomy poses a significant challenge in robot learning due to the need for high-precision, contact-rich dexterous manipulation involving delicate objects and soft tissue. Bimanual manipulation in surgery for handling delicate objects and transferring from one arm to another further adds to the complexity of the execution of surgical tasks. 
Learning dexterous skills on a surgical platform can be approached through learning-based algorithms to develop a library of policies for surgical tasks \cite{scheikl2023lapgym, xu2021surrol}. Reinforcement learning (RL) develops control policies by allowing an agent to iteratively interact with the surrounding environment to optimize its actions and maximize cumulative reward. RL approaches require extensive domain knowledge and reward engineering, hence, applying RL in surgical autonomy with safety-critical requirements may lead to undesirable, inappropriate, and non-robust actions by exploiting the physics and niches of the environment \cite{drolet2024comparison}.

\begin{figure}
    \centering
    \includegraphics[width=\linewidth]{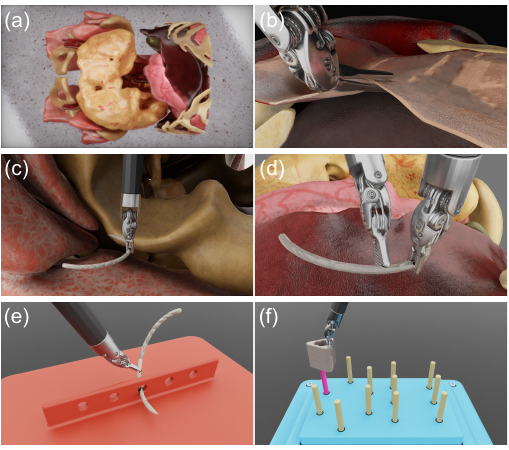}
    \caption{\textbf{Overview:} (a) Photorealistic human anatomical organs and textures in \simName. (b-f) Visuomotor behavior cloning policies executing fine-grained robotic maneuvers performed during surgery and the hands-on training exercises used in tabletop surgical curricula: (b) Tissue Retraction, (c) Needle Lift, (d) Needle Handover, (e) Suture Pad, (f) Block Transfer.
    }
    \label{fig:fig1}
\end{figure}

Behavior cloning (BC) provides an efficient approach to train policies from expert demonstrations. BC has been employed to perform a range of manipulation skills such as bimanual manipulation~\cite{drolet2024comparison}, dexterous manipulation~\cite{ding2024bunny}, and mobile manipulation \cite{fu2024mobile}. BC approaches require a vast number of expert demonstrations to be able to generalize to unseen, in-domain variations and novel objects \cite{chi2023diffusion, haldar2023teach}. Therefore, this necessitates access to an expensive and task-specific stack of surgical hardware and software to acquire the necessary demonstrations. As a result, progress in automated surgical subtasks has been slower.

In this work, we present \this: visual Behavior Cloning for Surgical First Interactive Autonomy Assistant to systematically evaluate the performance of state-of-the-art approaches in behavior cloning for augmented dexterity in RSAs. We extend \simName \cite{yu2024orbit} to provide an enhanced surgical digital twin simulator with photorealistic human anatomical organs and textures. This enables GPU-accelerated physics and ray-traced rendering in the context of photorealistic surgical scenes.

\newpage

\noindent Our primary contributions are as follows:
\begin{enumerate}
    \item We extend \simName to provide an enhanced surgical digital twin simulator with contact-rich physical interaction and high-fidelity visual rendering to facilitate the collection of synthetic surgical data for scaling data curation and teleoperation pipelines for surgical autonomy. 
    \item We create a set of photorealistic human anatomical organs and textures specifically designed for the surgical digital twin in \nvidia Omniverse.
    \item We demonstrate the effectiveness of behavior cloning approaches in solving fine-grained manipulation tasks for surgical augmented dexterity.
    \item We provide an open-source, high-quality teleoperation training dataset for all our experiments, collected by a human operator. This dataset includes tasks involving fine-grained dexterous manipulation of small and deformable objects that can be used for further learning and perception pipelines beyond this work.
    \item We systematically evaluate the generalization of our workflow across various surgical tasks.
\end{enumerate}

\section{Related Work}

\paragraph{\textbf{Behavior Cloning}}%
BC~\cite{Pomerleau2015ALVINNAA} is an imitation learning technique that learns a policy in a supervised manner by mapping observations to actions, using expert demonstrations. This approach has proven effective for various robotics applications \cite{chalodhorn2007learning, schaal2005learning, pastor2009learning} and manipulation tasks \cite{khansari2011learning, billard2004discovering, mulling2013learning}, where replicating expert behavior is crucial. However, traditional BC methods \cite{chalodhorn2007learning, florence2021implicitbehavioralcloning, ross2011reductionimitationlearningstructured} often struggle with tasks that require long-horizon planning or temporally extended behaviors, as these models tend to predict single-step actions without considering the broader action sequences needed to complete complex tasks. More recently, two lines of work have demonstrated improvements in BC by leveraging sequence modeling techniques to handle these challenges: Action Chunking with Transformers (ACT) \cite{zhao2023learningfinegrainedbimanualmanipulation} has excelled in generating coherent and precise actions across a range of complex and fine-grained manipulation tasks\cite{kim2024surgical, zhao2024aloha}, while Diffusion Policy \cite{chi2023diffusion} has shown promising results in various robotic tasks by effectively modeling expressive, multi-modal action distributions \cite{ze20243ddiffusionpolicygeneralizable}.

\paragraph{\textbf{Learning in Surgical Robotics}}
Surgical augmented dexterity involves human-supervised automation of minimal sub-tasks, enabling highly precise actuation. This approach has been applied to a range of sub-tasks with varying levels of autonomy, including dexterous needle picking and handling \cite{lin2023end, wilcox2022learning}, suturing \cite{joglekar2024autonomous, hari2024stitch, pedram2017autonomous}, and tissue manipulation \cite{li2020super, tagliabue2020soft}. \cite{alterovitz2003sensorless} performs numerical optimization with a soft tissue simulation based on a dynamic FEM formulation to model surgical steerable needle insertion. However, limited generalizability restricts the utility of these models in complex and unseen in-domain variations prevalent in the surgical domain. End-to-end imitation learning approaches for augmented dexterity have been explored in \cite{kim2024surgical, scheikl2024movement, li20223d}. Recent work by Kim et. al. \cite{kim2024surgical} introduces surgical policies that emphasize real-world capabilities, however, this approach is limited to specialized hardware access. Data-efficient imitation learning workflows with robust generalizability in complex surgical scenes remain elusive.

\paragraph{\textbf{Surgical Digital Twin}}
Surgical digital twins are real-time virtual models that replicate surgical environments or patient anatomies with high precision. They have become essential tools for enhancing surgical simulation \cite{yu2024orbit, scheikl2023lapgym, xu2021surrol}, planning \cite{moghani2024sufia}, and robotic surgery \cite{barnoy2021roboticsurgeryleanreinforcement, DATTA2021329}. These digital replicas enable the training of robot policies by offering realistic environments, reducing the sim-to-real gap, and helping their deployment in real-world surgical procedures.

\cite{bjelland2022toward} introduces a patient-specific digital twin concept for arthroscopic surgery, integrating soft tissue simulations, haptic feedback, and diagnostic data to improve surgical precision. \cite{shu2023twinsdigitaltwinskullbase} focuses on skull base surgery, where a high-precision digital twin system provides augmented surgical views in real-time. Most recently, \cite{hein2024creatingdigitaltwinspinal} showcases the dynamic 3D reconstruction of a complete surgical scene using multiple RGB-D cameras and data fusion techniques to provide high-fidelity scene reconstruction for proof-of-concept surgery digitalization.

\section{Problem Description}

We focus on learning visoumotor policies for fundamental surgical tasks and maneuvers from expert demonstrations for surgical augmented dexterity.

All of our experiments are conducted on the da Vinci Research Kit (dVRK) robot platform~\cite{kazanzides2014open} in the \simName~\cite{yu2024orbit} simulator. 5 photorealistic tasks representing fundamental surgical maneuvers are designed for this study. Each task has specific randomization with respect to the available workspace described in the simulation task section. The simulator provides access to RGB-D cameras with known intrinsic matrices, allowing for transformation between the camera's perspective and the world coordinate space. For point cloud down-sampling, we assume access to ground-truth semantic instance segmentation masks in simulation.

\section{Method}

\subsection{Building Surgical Digital Twins}

In the rapidly evolving field of surgical simulation, the quality and realism of 3D anatomical models play a crucial role in providing effective training experiences for surgeons \cite{SARAMIS_Montana_Brown}. These models serve as the foundation for virtual surgical environments, allowing medical professionals to practice complex procedures in a risk-free setting that closely mimics real-world scenarios. Our innovative approach leverages cutting-edge AI technology and advanced 3D modeling techniques to produce anatomically accurate and visually compelling models for surgical simulation purposes.

Our pipeline for 3D modelling has two scan sources: synthetic organ generation using NVIDIA MAISI \cite{MONAI_MAISI_2024} and segmentation from real CTs using VISTA3D or Auto3DSeg \cite{DIAZPINTO_MONAI}. MAISI generates high-resolution synthetic CT images and corresponding segmentation masks with up to 127 anatomical classes, achieving landmark voxel dimensions of 512 x 512 × 512 and spacing of 1.0 × 1.0 × 1.0 $mm^3$. To streamline the process, we group the segmented structures into 17 broader anatomical regions. For instance, instead of individual vertebrae, we create a unified "spine" group. Similarly, individual ribs are consolidated into a "ribs" group. This approach is applied consistently across muscles, lungs, blood vessels, and hip structures, creating a more manageable set of anatomical units for further processing.

Following AI-assisted segmentation and synthesis, anatomical structures are converted into 3D mesh models using the marching cubes algorithm \cite{Lorensen_Marching_cubes}. These models then undergo extensive refinement using advanced 3D modeling tools, including dynamic remeshing, topology optimization, strategic use of masking techniques, and generation of high-quality UV maps. This process is meticulously applied to all anatomical regions and individual organs. While time-consuming, this manual refinement is crucial for achieving clean meshes and proper UV maps necessary for realistic materials and deformable body simulations.

The refined models undergo a comprehensive texturing process, employing physically based rendering materials, subsurface scattering, and custom layers for specific anatomical details. A custom shader is then created in \nvidia Omniverse to accurately represent the material properties of each anatomical structure. The textured and shader-equipped models are assembled into a unified OpenUSD (Universal Scene Description) file, ready for integration with various surgical simulation platforms.

This pipeline offers several advantages for surgical simulation, including enhanced realism that contributes to immersive training experiences and high anatomical accuracy derived from AI-segmented medical imaging data. Additionally, the workflow allows for the creation of patient-specific models, enabling the simulation of rare or complex cases. The use of standard 3D formats ensures multi-platform compatibility, while the detailed geometry and accurate materials provide a solid foundation for implementing realistic haptic feedback.

\begin{figure}
    \centering
    \includegraphics[width=0.49\textwidth]{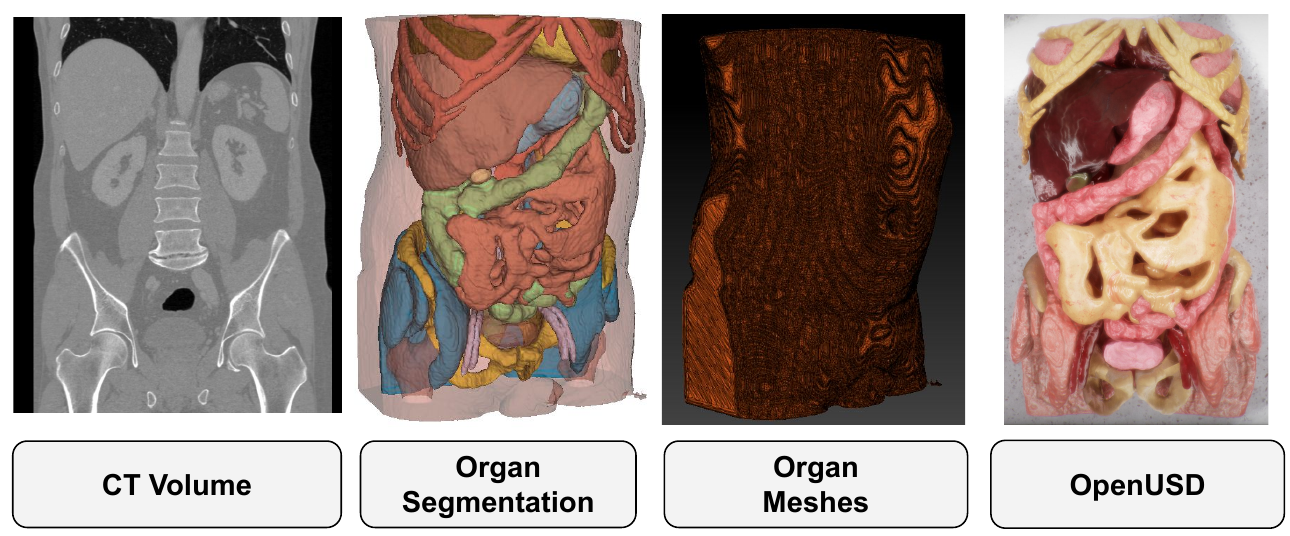}
    \caption{\textbf{Surgical digital twin:} This workflow illustrates the full pipeline for creating photorealistic anatomical models, from raw CT volume data to final OpenUSD in \nvidia Omniverse. The process includes organ segmentation, mesh conversion, mesh cleaning and refinement, photorealistic texturing, and culminating in the assembly of all textured organs into a unified OpenUSD file.
    }
    \label{fig:fig2}
\end{figure}

\subsection{Imitation Learning}

Given a set of expert demonstrations collected through teleoperation, we learn a visuomotor policy $\pi: \mathcal{O} \rightarrow \mathcal{A}$ that maps observations $\mathbf{o} \in \mathcal{O}$, that include RGB-D images and proprioceptive data, to actions $\mathbf{a} \in \mathcal{A}$. State-of-the-art behavior cloning (BC) decision backbones with robust visual representations enable learning surgical policies capable of performing high-precision dual-arm tasks.

\paragraph{\textbf{Perception}} Our method investigates two types of visual features: (1) those derived from raw RGB images, and (2) those from sparse point clouds generated using depth information. We utilize RGB-D images sourced from either a third-person or an endoscopic camera, as input for visual observations. For image-based features, we train a ResNet-18 for each camera view, producing a view-specific feature embedding. For point cloud features, we generate sparse point clouds from the depth data, using the camera’s intrinsic parameters for projection. To focus on task-relevant objects, we apply segmentation or boundary cropping and exclude color channels to filter out redundant information. Farthest point sampling \cite{qi2017pointnet} is used for downsampling to obtain a sparse 3D point cloud.

\paragraph{\textbf{Policy Learning with Behavior Cloning}}
To handle decision-making, we utilize two state-of-the-art decision backbones:

\noindent \textbf{Action Chunking with Transformers (ACT):} ACT \cite{zhao2023learningfinegrainedbimanualmanipulation} trains a Conditional Variational Autoencoder (CVAE) within a multi-headed attention transformer framework. ACT predicts sequences of actions, referred to as action chunks, using a history of observations. To improve the smoothness and consistency of multi-step action generation, ACT utilizes temporal aggregation, averaging overlapping action chunks across timesteps. This allows ACT to model temporally extended behavior, making it particularly effective for tasks that require coordinated, multi-step actions \cite{zhao2024aloha}.

\noindent \textbf{Diffusion Policy:} \cite{chi2023diffusion} utilizes expressive generative models \cite{ho2020denoisingdiffusionprobabilisticmodels} for behavior cloning by gradually refining action sequences through a series of learned transitions. Instead of predicting a single deterministic action or a chunk of actions, diffusion policies start with a noisy sequence of actions and iteratively denoise them to generate coherent sequences. This iterative process allows diffusion policies to better capture the uncertainty and multimodality common in robotic tasks.

The visual feature embeddings extracted by the perception module $\mathbf{z}$, are concatenated with proprioceptive data $\mathbf{q}$, which consists of joint positions and gripper states. This combined input is then passed to the backbone to generate relative end-effector robot actions $\mathbf{a} = \pi(\mathbf{z}, \mathbf{q})$. We investigate the impact of each vision and controller backbone on learning-based surgical policy performance and conduct thorough ablation studies to assess their respective strengths in handling different surgical tasks.

\section{Experimental Results}

\subsection{Experiment Setup}

\begin{figure}[!t]
    \centering
    \includegraphics[width=0.95\linewidth]{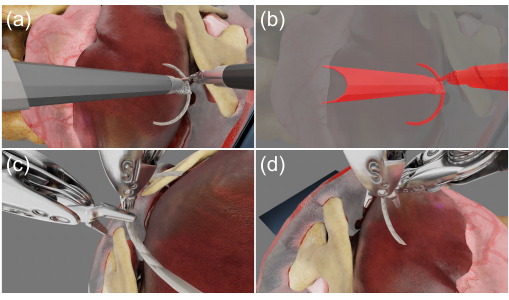}
    \caption{\textbf{Visual observations}: An example of visual observations for the needle handover; (a) a static top down endoscope view, (b) downsampled point clouds capturing the arms and the suture needle -- the background is only shown for demonstration purposes and is not used during training for point cloud-based policies, (c-d) wrist camera views from each arm.
    }
    \label{fig:fig3}
\end{figure}

\begin{table*}[]
\centering
\begin{minipage}{0.66\textwidth}
\centering
\resizebox{\linewidth}{!}{%
\begin{tabular}{l|c|c|c|c}
\toprule
\rowcolor[HTML]{CBCEFB} 
Task & ACT & ACT  & ACT  & DP3 \\
\rowcolor[HTML]{CBCEFB} 
 & Single Camera & Multi Camera & Point Cloud & Point Cloud \\
\midrule
Tissue Retraction & 0.98 & \textbf{1.00} & 0.58 & 0.80 \\
\rowcolor[HTML]{EFEFEF} 
Needle Lift & 0.68 & 0.70 & 0.70 & \textbf{0.88} \\
Needle Handover & 0.14 & 0.42 & 0.34 & \textbf{0.58} \\
\rowcolor[HTML]{EFEFEF} 
Suture Pad & \textbf{0.48} & 0.40 & 0.36 & 0.46 \\
Block Transfer & \textbf{0.78} & 0.46 & 0.68 & 0.30 \\
\bottomrule
\end{tabular}%
}
\end{minipage}
\hspace{10pt}
\begin{minipage}{0.23\textwidth}
\centering
\caption{\textbf{Simulation results:} Success rates for surgical tasks using 3D Diffusion Policy (DP3) and Action Chunking Transformers (ACT) across different vision modalities are evaluated over 50 test trials. Results highlight trade-offs among models across tasks with unique requirements.}
\label{tab:table1}
\end{minipage}
\end{table*}

Our simulation experiments are carried out in \simName~\cite{yu2024orbit} which precisely simulates joint articulation and low-level controllers of the physical dVRK platform, supports high-precision, contact-rich physical interactions, and delivers high-fidelity realistic rendering. In our experiments, we utilize camera sensors to acquire $512 \times 512$ RGB-D images. An example of visual observations for policy learning is illustrated in Fig. \ref{fig:fig3}. The wrist camera for surgical manipulation was introduced in \cite{kim2024surgical}. In our experiments, we explore three observation spaces; "Single Camera", "Multi Camera", and "Point Cloud". All include joint position information and only differ in the vision modality. "Single Camera" employs a primary camera view that includes a static endoscopic top-down view for the tasks involving suture needle and tissue retraction, and a side view camera to acquire visual input for the tabletop block transfer and suture pad tasks. "Multi Camera" includes a wrist camera in addition to the primary task camera. "Point Cloud" utilizes segmented point cloud derived from the primary task camera.

\subsection{Expert Demonstrations with Teleoperation}

For each task, a dataset of 50 expert demonstrations is collected through human teleoperation across all tasks. The demonstrations are recorded using virtual reality (VR) controllers, which provide six-degree-of-freedom (6-DoF) positional and rotational input, enabling precise control of the dVRK gripper in Cartesian space.

The choice of VR controllers offers several advantages. VR controllers are widely available and cost-effective compared to specialized surgical consoles or haptic devices, making them accessible to the broader research community. Second, they provide an intuitive interface for teleoperation, facilitating the collection of high-quality demonstrations without the need for extensive operator training.

\subsection{Implementation Details}

We adopt the original ACT architecture \cite{zhao2023learningfinegrainedbimanualmanipulation} without modifications for image-based experiments. To integrate point clouds into ACT, we follow the method in \cite{zhu2024pointcloudmattersrethinking}, using PointNet \cite{qi2017pointnet} as the point cloud encoder. Positional embeddings are added to the per-point features from PointNet, which are concatenated with joint information and a style variable, and then passed to the CVAE decoder for action prediction.

We utilize 3D Diffusion Policy (DP3) \cite{ze20243ddiffusionpolicygeneralizable} to explore point cloud-based diffusion models, training exclusively on depth information with a U-Net backbone, and predict the final sample directly, rather than predicting noise (epsilon).

\subsection{Simulation Tasks and Evaluation Metrics}

The experimental environments are shown in Fig. \ref{fig:fig1}. The tasks are chosen to represent both dexterous in-vivo robotic maneuvers performed during surgery and hands-on training exercises used in tabletop surgical curricula \cite{fls}. The task description and their success metrics are outlined below:

\texttt{Tissue Retraction} -- A piece of soft skin tissue is placed on top of a liver organ. A point with a uniform random distribution of $4$ cm along the edge of the tissue is marked to indicate the desired robot grasping region. The robot grasps the tissue with forceps grippers and lifts it upwards. The task is considered successful when the robot correctly grasps the tissue at the specified location and lifts it up.

\texttt{Needle Lift} -- A suture needle is initialized with a uniform distribution of $\pm 2.5$ cm in the x-axis and $\pm 1$ cm on the y-axis on the available space of the back muscle organ. The task is successful if the robot grasps and lifts the needle to a specified height.

\texttt{Needle Handover} -- This task involves a dual-arm dVRK setup. A suture needle is initialized with a uniform distribution of $\pm 1.5$ cm in the x-axis and $\pm 2$ cm on the y-axis on the available space of the liver organ. The right arm first grasps the needle and transfers it to a handover location. The left arm then grasps the needle from the handover point and takes it away. The task is considered successful when the needle is successfully transferred from one arm to the other.

\texttt{Suture Pad} -- A suture needle is initialized with a uniform distribution of $\pm 2$ cm along the top of a suture pad. The robot first grasps the needle and threads it through the center hole of the suture pad. The task is considered successful when the needle is threaded through the hole to the opposite side.

\texttt{Block Transfer} -- A surgical training block is randomly initialised on a peg tower board. The dVRK arm grasps the block, properly lifts it above the tower, and then transfers it to a designated pink tower. The task is successful if the robot moves the block from initial to the goal tower.

\begin{figure*}
    \centering
    \includegraphics[width=\textwidth]{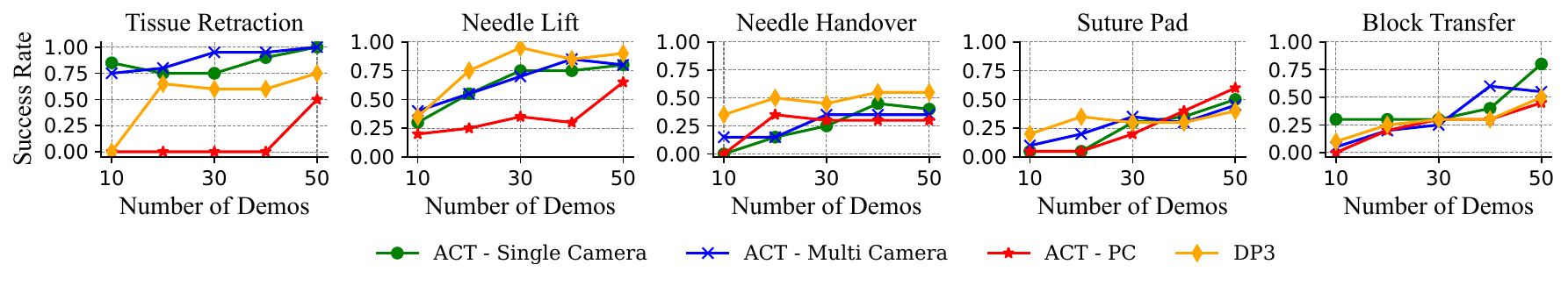}
    \caption{\textbf{Sample efficiency in simulation:} Each task's 50 demonstrations are subsampled in increments of 10 to evaluate the models' sample efficiency. Success rates are calculated over 20 trial runs at test time.
    }
    \label{fig:fig4}
\end{figure*}

\subsection{Evaluation}

Our main experimental results are reported in TABLE \ref{tab:table1}. We evaluate a range of surgical tasks, including needle lift, tissue retraction, needle handover, suture pad threading, and block transfer. We observe that while all policies perform well when joint position information predominates, tasks requiring enhanced perceptual precision show different behavior based on the perception encoder.

Most models achieve high success rates for the needle lift and tissue retraction tasks which involve straightforward actions, i.e. lifting or retracting objects. These results suggest that when tasks primarily depend on accurate joint positioning, the common state encoder is sufficient, reducing the reliance on high-precision grasping. In tissue retraction, RGB-based models perform slightly better, likely due to a distinct red marker that clearly indicates a visual cue for the grasping point. In contrast, point cloud encoders, which lack color information, must detect the grasping point from a sparse set of points that blend with the surrounding tissue.

For tasks such as needle handover, suture pad threading, and block transfer, which require high-precision and contact-rich interactions, performance is heavily dependent on accurate grasping of small objects and precise manipulation to the goal pose. As a result, successful manipulation imposes greater demands on the visual perception module, requiring it to accurately ground small surgical objects in the scene and effectively account for the characteristics of their slender geometry.

Point cloud-based policies tend to outperform their RGB-based counterparts in tasks where manipulator-object interactions are captured more explicitly by their spatial coordinates. For example, the slender design of the dVRK arm and the needle's perpendicular orientation relative to the arm enable point cloud representations to capture precise spatial relationships between the manipulator and the object.

An exception within this category is the block transfer task, where the single-camera RGB model demonstrates relatively strong performance. This outcome can be explained by the constrained variability in the task setup: although block placement is randomized, it is limited to six pegs, and the goal location remains constant throughout training and evaluation. In contrast, point cloud-based models exhibit comparatively poorer performance. The absence of color information makes it challenging for these models to differentiate objects with similar shapes and orientations—such as the arm, block, and individual pegs—in cluttered scenes, especially when the arm is positioned close to objects intended for grasping. This limitation underscores the importance of color and texture cues in tasks involving objects with subtle geometric differences.

\subsection{Sample Efficiency}

Expert demonstrations are both difficult to obtain and limited in quantity due to the specialized skills required and the high cost of data collection. As a result, sample efficiency becomes critical, as we aim for policies to learn effectively with minimal demonstrations. We evaluate the performance of both ACT and diffusion policies across varying numbers of demonstrations used for training and present the results in Fig. \ref{fig:fig4}.

We observe several common failure modes across all models, which highlight challenges in sample efficiency. With fewer demonstrations (10-20), failures often involve successful object identification but incomplete task execution, such as grasping without lifting or slight misalignment, with overfitting to narrow trajectories limiting adaptability. As training increases (30-40 demos), failures shift to issues with precise grasps and the absence of corrective actions when small errors occur, such as continuing a motion despite missing the object. Even with 50 demonstrations, persistent issues including inaccurate object positioning or grasp instability remain. These observations emphasize the need for policies that are not only sample-efficient but also capable of adaptive and corrective behavior.

\begin{figure*}
    \centering
    \includegraphics[width=0.95\textwidth]{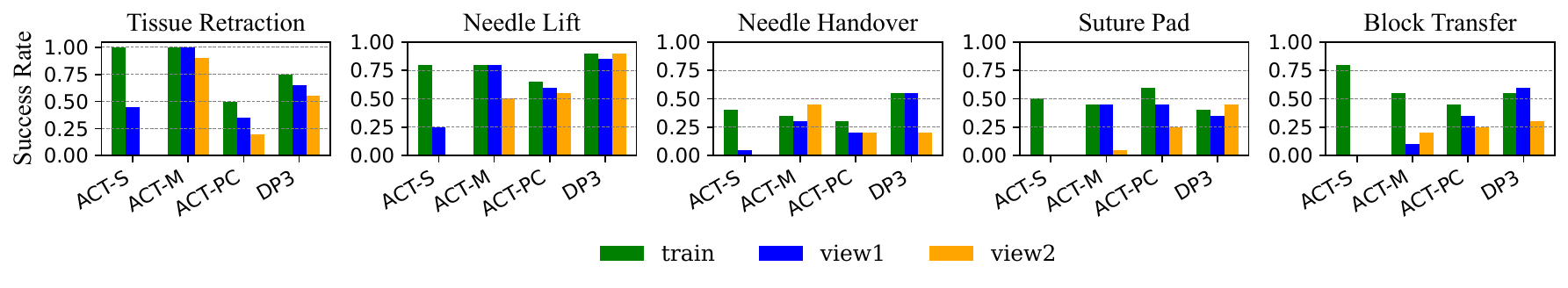}
    \caption{\textbf{Viewpoint robustness:} Performance of models trained on the primary camera views (train) is evaluated against small camera perturbations (view 1) and major viewpoint changes (view 2). Success rates are calculated over 20 trial runs at test time. ACT - S, ACT - M, and ACT - PC denote ACT - Single Camera, ACT - Multi Camera, and ACT - Point Cloud, respectively.
    }
    \label{fig:fig5}
\end{figure*}

\subsection{Instance Generalization}

To achieve practical deployment in real physical experiments, policies must exhibit generalization capabilities, particularly in handling variations in task-relevant tools such as surgical needles. We evaluate the robustness of the trained policies for lifting the primary suture needle instance (Needle N1 shown in TABLE \ref{tab:table2}) by reporting their success on four out-of-distribution previously unseen needle instances of varying sizes and irregular shapes (Needle N2 - N5 shown in TABLE \ref{tab:table2}).

Our results show that point cloud-based models exhibit poorer generalization to new needle instances, performing noticeably worse than the RGB models. We hypothesize that this is due to overfitting to the specific geometry of the training needle (N1), as the depth information provides a more detailed geometric representation. Qualitatively, these models often fail to localize the needle accurately, sometimes even moving the gripper in the opposite direction. We also observe overfitting to the size of N1, which is larger than needles N2 and N3. As a result, the point cloud models frequently attempt to grasp slightly above the smaller needles, leading to missed attempts.

In contrast, image-based models demonstrate better generalization across needle instances, likely due to their reliance on color and texture information, which provides more robust visual cues and less sensitivity to geometric details than depth-based models. The multi-camera setup further enhances generalization by providing diverse perspectives, which appears to help the model form a more robust understanding of the target object.

\begin{table}[]
\centering
\includegraphics{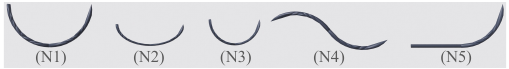}
\resizebox{\columnwidth}{!}{%
\begin{tabular}{l|c|c|c|c}
\toprule
\rowcolor[HTML]{CBCEFB} 
Test Needle & \begin{tabular}[c]{@{}c@{}}ACT\\ Single Camera\end{tabular} & \begin{tabular}[c]{@{}c@{}}ACT\\ Multi Camera\end{tabular} & \begin{tabular}[c]{@{}c@{}}ACT\\ Point Cloud\end{tabular} & DP3 \\
\midrule
\midrule
N1 (train) & 0.68 & 0.70 & 0.70 & 0.88\\
\rowcolor[HTML]{EFEFEF} 
N2 & 0.24 & \textbf{0.44} & 0.26 & 0.18\\
N3 & 0.32 & \textbf{0.36} & 0.22 & 0.24\\
\rowcolor[HTML]{EFEFEF} 
N4 & 0.38 & \textbf{0.56} & 0.24 & 0.16\\
N5 & 0.30 & \textbf{0.68} & 0.08 & 0.14\\
\midrule
\rowcolor[HTML]{EFEFEF}
Average  & 0.31 & \textbf{0.51} & 0.20 & 0.18\\
\bottomrule
\end{tabular}%
}
\caption{\textbf{Instance generalization in simulation:} We assess the effectiveness of the policies trained only on the primary suture needle (Needle N1) and report the success rates and averages for lifting previously unseen suture needles with irregular shapes (Needles N2 - N5) over 50 trial runs at test time.
}
\label{tab:table2}
\end{table}

\vspace{-0.05in}
\subsection{Viewpoint Robustness}

To bridge the sim-to-real gap, surgical policies must be robust to changes in camera position and orientation, as replicating exact setups in real-world environments is challenging. This issue is paramount in surgical tasks where precision is critical. We evaluate this robustness by conducting two experiments with varying camera viewpoint alterations. In the first experiment (view 1 in Fig. \ref{fig:fig5}), we introduce small shifts \(\pm 1cm\) in translation and \(\pm 5\)\textdegree \ rotation in all axis to the camera’s original training position. In the second experiment (view 2 in Fig. \ref{fig:fig5}), we introduce a more drastic change by completely swapping camera perspectives — policies trained with an endoscopic view are evaluated with a third-person view, and vice versa. This allows us to measure resilience to both minor and major viewpoint shifts.

In experiments with small camera perturbation, ACT-PC, DP3, and ACT-M show strong robustness, with minimal performance degradation. In contrast, ACT-S experienced a significant drop. For experiments with a major viewpoint shift, both RGB models (ACT-S and ACT-M) fail, with success rates significantly dropping across all tasks. While point cloud-based models also decline, the drop is much smaller. Tasks like needle lift, suture, and tissue retraction remain relatively robust, highlighting the superior resilience of point cloud models to both minor and major camera perspective changes. Additionally, while ACT-M demonstrate the greatest robustness of the RGB-based models, it is important to note that the wrist-mounted camera positions did not change across experiments.

\vspace{-0.05in}

\section{Limitations}

It is important to note that the fidelity of digital twin models is ultimately limited by the information derived from CT scans. Connective tissue, in particular, remains a challenging aspect to model accurately due to its complex and varied nature in medical imaging. Despite these limitations, the photorealistic models produced by this pipeline have the potential to improve surgical training and enhance skill transfer to real-world procedures.

Our experiments are carried out in scenes that are consistent throughout training and evaluation. While all training and evaluation are completed with objects initialized using uniform randomization, further work is needed to enhance the generalization performance of surgical robotic policies.

\section{Conclusion}

In this work, we introduced \this to advance research on imitation learning-based assistive autonomy in robotic surgery. We find that the diverse set of tasks and surgical scenes introduced in our framework enables an in-depth analysis of the capabilities and limitations of current state-of-the-art imitation learning policies and their core components. By examining performance across manipulation tasks of varying complexity and dataset sizes, we find that current RGB-based models offer rich semantic understanding critical to surgical scenes, but require multiple cameras to offer greater instance and viewpoint robustness. Additionally, we observe that point cloud-based models offer excellent viewpoint robustness, with the caveat that they tend to be sensitive to target object geometries. These insights inform the development of more effective surgical robotic policies, emphasizing the need to further develop perception representations that balance the strengths of the state-of-the-art policies evaluated in this work.

\clearpage

\bibliographystyle{IEEEtran}
\bibliography{root}

\begin{thebibliography}{10}
\providecommand{\url}[1]{#1}
\csname url@samestyle\endcsname
\providecommand{\newblock}{\relax}
\providecommand{\bibinfo}[2]{#2}
\providecommand{\BIBentrySTDinterwordspacing}{\spaceskip=0pt\relax}
\providecommand{\BIBentryALTinterwordstretchfactor}{4}
\providecommand{\BIBentryALTinterwordspacing}{\spaceskip=\fontdimen2\font plus
\BIBentryALTinterwordstretchfactor\fontdimen3\font minus \fontdimen4\font\relax}
\providecommand{\BIBforeignlanguage}[2]{{%
\expandafter\ifx\csname l@#1\endcsname\relax
\typeout{** WARNING: IEEEtran.bst: No hyphenation pattern has been}%
\typeout{** loaded for the language `#1'. Using the pattern for}%
\typeout{** the default language instead.}%
\else
\language=\csname l@#1\endcsname
\fi
#2}}
\providecommand{\BIBdecl}{\relax}
\BIBdecl

\bibitem{kyg}
K.~Goldberg and G.~Guthart, ``Augmented dexterity: How robots can enhance human surgical skills,'' \emph{Science Robotics}, vol.~9, no.~95, p. eadr5247, 2024.

\bibitem{scheikl2023lapgym}
P.~M. Scheikl, B.~Gyenes, R.~Younis, C.~Haas, G.~Neumann, M.~Wagner, and F.~Mathis-Ullrich, ``{LapGym} - an open source framework for reinforcement learning in robot-assisted laparoscopic surgery,'' \emph{Journal of Machine Learning Research}, vol.~24, no. 368, pp. 1--42, 2023.

\bibitem{xu2021surrol}
J.~Xu, B.~Li, B.~Lu, Y.-H. Liu, Q.~Dou, and P.-A. Heng, ``Surrol: An open-source reinforcement learning centered and dvrk compatible platform for surgical robot learning,'' in \emph{2021 IEEE/RSJ International Conference on Intelligent Robots and Systems (IROS)}.\hskip 1em plus 0.5em minus 0.4em\relax IEEE, 2021, pp. 1821--1828.

\bibitem{drolet2024comparison}
M.~Drolet, S.~Stepputtis, S.~Kailas, A.~Jain, J.~Peters, S.~Schaal, and H.~B. Amor, ``A comparison of imitation learning algorithms for bimanual manipulation,'' \emph{IEEE Robotics and Automation Letters}, 2024.

\bibitem{ding2024bunny}
R.~Ding, Y.~Qin, J.~Zhu, C.~Jia, S.~Yang, R.~Yang, X.~Qi, and X.~Wang, ``Bunny-visionpro: Real-time bimanual dexterous teleoperation for imitation learning,'' \emph{arXiv preprint arXiv:2407.03162}, 2024.

\bibitem{fu2024mobile}
Z.~Fu, T.~Z. Zhao, and C.~Finn, ``{Mobile ALOHA}: Learning bimanual mobile manipulation with low-cost whole-body teleoperation,'' \emph{arXiv preprint arXiv:2401.02117}, 2024.

\bibitem{chi2023diffusion}
C.~Chi, S.~Feng, Y.~Du, Z.~Xu, E.~Cousineau, B.~Burchfiel, and S.~Song, ``Diffusion policy: Visuomotor policy learning via action diffusion,'' \emph{arXiv preprint arXiv:2303.04137}, 2023.

\bibitem{haldar2023teach}
S.~Haldar, J.~Pari, A.~Rai, and L.~Pinto, ``Teach a robot to fish: Versatile imitation from one minute of demonstrations,'' \emph{arXiv preprint arXiv:2303.01497}, 2023.

\bibitem{yu2024orbit}
Q.~Yu, M.~Moghani, K.~Dharmarajan, V.~Schorp, W.~C.-H. Panitch, J.~Liu, K.~Hari, H.~Huang, M.~Mittal, K.~Goldberg, and A.~Garg, ``{ORBIT-Surgical}: An open-simulation framework for learning surgical augmented dexterity,'' \emph{arXiv preprint arXiv:2404.16027}, 2024.

\bibitem{Pomerleau2015ALVINNAA}
D.~A. Pomerleau, ``Alvinn: An autonomous land vehicle in a neural network,'' \emph{Advances in neural information processing systems}, vol.~1, 1988.

\bibitem{chalodhorn2007learning}
R.~Chalodhorn, D.~B. Grimes, K.~Grochow, and R.~P. Rao, ``Learning to walk through imitation.'' in \emph{IJCAI}, vol.~7, 2007, pp. 2084--2090.

\bibitem{schaal2005learning}
S.~Schaal, J.~Peters, J.~Nakanishi, and A.~Ijspeert, ``Learning movement primitives,'' in \emph{Robotics Research. The Eleventh International Symposium: With 303 Figures}.\hskip 1em plus 0.5em minus 0.4em\relax Springer, 2005, pp. 561--572.

\bibitem{pastor2009learning}
P.~Pastor, H.~Hoffmann, T.~Asfour, and S.~Schaal, ``Learning and generalization of motor skills by learning from demonstration,'' in \emph{2009 IEEE International Conference on Robotics and Automation}.\hskip 1em plus 0.5em minus 0.4em\relax IEEE, 2009, pp. 763--768.

\bibitem{khansari2011learning}
S.~M. Khansari-Zadeh and A.~Billard, ``Learning stable nonlinear dynamical systems with gaussian mixture models,'' \emph{IEEE Transactions on Robotics}, vol.~27, no.~5, pp. 943--957, 2011.

\bibitem{billard2004discovering}
A.~Billard, Y.~Epars, S.~Calinon, S.~Schaal, and G.~Cheng, ``Discovering optimal imitation strategies,'' \emph{Robotics and autonomous systems}, vol.~47, no. 2-3, pp. 69--77, 2004.

\bibitem{mulling2013learning}
K.~M{\"u}lling, J.~Kober, O.~Kroemer, and J.~Peters, ``Learning to select and generalize striking movements in robot table tennis,'' \emph{The International Journal of Robotics Research}, vol.~32, no.~3, pp. 263--279, 2013.

\bibitem{florence2021implicitbehavioralcloning}
P.~Florence, C.~Lynch, A.~Zeng, O.~A. Ramirez, A.~Wahid, L.~Downs, A.~Wong, J.~Lee, I.~Mordatch, and J.~Tompson, ``Implicit behavioral cloning,'' in \emph{Conference on Robot Learning}.\hskip 1em plus 0.5em minus 0.4em\relax PMLR, 2022, pp. 158--168.

\bibitem{ross2011reductionimitationlearningstructured}
S.~Ross, G.~Gordon, and D.~Bagnell, ``A reduction of imitation learning and structured prediction to no-regret online learning,'' in \emph{Proceedings of the fourteenth international conference on artificial intelligence and statistics}.\hskip 1em plus 0.5em minus 0.4em\relax JMLR Workshop and Conference Proceedings, 2011, pp. 627--635.

\bibitem{zhao2023learningfinegrainedbimanualmanipulation}
T.~Z. Zhao, V.~Kumar, S.~Levine, and C.~Finn, ``Learning fine-grained bimanual manipulation with low-cost hardware,'' \emph{arXiv preprint arXiv:2304.13705}, 2023.

\bibitem{kim2024surgical}
J.~W. Kim, T.~Z. Zhao, S.~Schmidgall, A.~Deguet, M.~Kobilarov, C.~Finn, and A.~Krieger, ``{Surgical Robot Transformer (SRT)}: Imitation learning for surgical tasks,'' \emph{arXiv preprint arXiv:2407.12998}, 2024.

\bibitem{zhao2024aloha}
T.~Z. Zhao, J.~Tompson, D.~Driess, P.~Florence, S.~K.~S. Ghasemipour, C.~Finn, and A.~Wahid, ``{ALOHA Unleashed}: A simple recipe for robot dexterity,'' in \emph{8th Annual Conference on Robot Learning}.

\bibitem{ze20243ddiffusionpolicygeneralizable}
Y.~Ze, G.~Zhang, K.~Zhang, C.~Hu, M.~Wang, and H.~Xu, ``{3D} diffusion policy: Generalizable visuomotor policy learning via simple 3d representations,'' in \emph{ICRA 2024 Workshop on 3D Visual Representations for Robot Manipulation}, 2024.

\bibitem{lin2023end}
H.~Lin, B.~Li, X.~Chu, Q.~Dou, Y.~Liu, and K.~W.~S. Au, ``End-to-end learning of deep visuomotor policy for needle picking,'' in \emph{2023 IEEE/RSJ International Conference on Intelligent Robots and Systems (IROS)}.\hskip 1em plus 0.5em minus 0.4em\relax IEEE, 2023, pp. 8487--8494.

\bibitem{wilcox2022learning}
A.~Wilcox, J.~Kerr, B.~Thananjeyan, J.~Ichnowski, M.~Hwang, S.~Paradis, D.~Fer, and K.~Goldberg, ``Learning to localize, grasp, and hand over unmodified surgical needles,'' in \emph{2022 International Conference on Robotics and Automation (ICRA)}.\hskip 1em plus 0.5em minus 0.4em\relax IEEE, 2022, pp. 9637--9643.

\bibitem{joglekar2024autonomous}
N.~Joglekar, F.~Liu, F.~Richter, and M.~C. Yip, ``Autonomous image-to-grasp robotic suturing using reliability-driven suture thread reconstruction,'' \emph{arXiv preprint arXiv:2408.16938}, 2024.

\bibitem{hari2024stitch}
K.~Hari, H.~Kim, W.~Panitch, K.~Srinivas, V.~Schorp, K.~Dharmarajan, S.~Ganti, T.~Sadjadpour, and K.~Goldberg, ``{STITCH}: Augmented dexterity for suture throws including thread coordination and handoffs,'' \emph{arXiv preprint arXiv:2404.05151}, 2024.

\bibitem{pedram2017autonomous}
S.~A. Pedram, P.~Ferguson, J.~Ma, E.~Dutson, and J.~Rosen, ``Autonomous suturing via surgical robot: An algorithm for optimal selection of needle diameter, shape, and path,'' in \emph{2017 IEEE International conference on robotics and automation (ICRA)}.\hskip 1em plus 0.5em minus 0.4em\relax IEEE, 2017, pp. 2391--2398.

\bibitem{li2020super}
Y.~Li, F.~Richter, J.~Lu, E.~K. Funk, R.~K. Orosco, J.~Zhu, and M.~C. Yip, ``Super: A surgical perception framework for endoscopic tissue manipulation with surgical robotics,'' \emph{IEEE Robotics and Automation Letters}, vol.~5, no.~2, pp. 2294--2301, 2020.

\bibitem{tagliabue2020soft}
E.~Tagliabue, A.~Pore, D.~Dall’Alba, E.~Magnabosco, M.~Piccinelli, and P.~Fiorini, ``Soft tissue simulation environment to learn manipulation tasks in autonomous robotic surgery,'' in \emph{2020 IEEE/RSJ International Conference on Intelligent Robots and Systems (IROS)}.\hskip 1em plus 0.5em minus 0.4em\relax IEEE, 2020, pp. 3261--3266.

\bibitem{alterovitz2003sensorless}
R.~Alterovitz, K.~Goldberg, J.~Pouliot, R.~Taschereau, and I.-C. Hsu, ``Sensorless planning for medical needle insertion procedures,'' in \emph{Proceedings 2003 IEEE/RSJ International Conference on Intelligent Robots and Systems (IROS 2003)(Cat. No. 03CH37453)}, vol.~4.\hskip 1em plus 0.5em minus 0.4em\relax IEEE, 2003, pp. 3337--3343.

\bibitem{scheikl2024movement}
P.~M. Scheikl, N.~Schreiber, C.~Haas, N.~Freymuth, G.~Neumann, R.~Lioutikov, and F.~Mathis-Ullrich, ``Movement primitive diffusion: Learning gentle robotic manipulation of deformable objects,'' \emph{IEEE Robotics and Automation Letters}, 2024.

\bibitem{li20223d}
B.~Li, R.~Wei, J.~Xu, B.~Lu, C.~H. Yee, C.~F. Ng, P.-A. Heng, Q.~Dou, and Y.-H. Liu, ``{3D} perception based imitation learning under limited demonstration for laparoscope control in robotic surgery,'' in \emph{2022 International Conference on Robotics and Automation (ICRA)}.\hskip 1em plus 0.5em minus 0.4em\relax IEEE, 2022, pp. 7664--7670.

\bibitem{moghani2024sufia}
M.~Moghani, L.~Doorenbos, W.~C.-H. Panitch, S.~Huver, M.~Azizian, K.~Goldberg, and A.~Garg, ``{SuFIA}: Language-guided augmented dexterity for robotic surgical assistants,'' \emph{arXiv preprint arXiv:2405.05226}, 2024.

\bibitem{barnoy2021roboticsurgeryleanreinforcement}
Y.~Barnoy, M.~O'Brien, W.~Wang, and G.~Hager, ``Robotic surgery with lean reinforcement learning,'' \emph{arXiv preprint arXiv:2105.01006}, 2021.

\bibitem{DATTA2021329}
S.~Datta, Y.~Li, M.~M. Ruppert, Y.~Ren, B.~Shickel, T.~Ozrazgat-Baslanti, P.~Rashidi, and A.~Bihorac, ``Reinforcement learning in surgery,'' \emph{Surgery}, vol. 170, no.~1, pp. 329--332, 2021.

\bibitem{bjelland2022toward}
{\O}.~Bjelland, B.~Rasheed, H.~G. Schaathun, M.~D. Pedersen, M.~Steinert, A.~I. Hellevik, and R.~T. Bye, ``Toward a digital twin for arthroscopic knee surgery: a systematic review,'' \emph{IEEE Access}, vol.~10, pp. 45\,029--45\,052, 2022.

\bibitem{shu2023twinsdigitaltwinskullbase}
H.~Shu, R.~Liang, Z.~Li, A.~Goodridge, X.~Zhang, H.~Ding, N.~Nagururu, M.~Sahu, F.~X. Creighton, R.~H. Taylor \emph{et~al.}, ``{Twin-S}: A digital twin for skull base surgery,'' \emph{International journal of computer assisted radiology and surgery}, vol.~18, no.~6, pp. 1077--1084, 2023.

\bibitem{hein2024creatingdigitaltwinspinal}
J.~Hein, F.~Giraud, L.~Calvet, A.~Schwarz, N.~A. Cavalcanti, S.~Prokudin, M.~Farshad, S.~Tang, M.~Pollefeys, F.~Carrillo \emph{et~al.}, ``Creating a digital twin of spinal surgery: A proof of concept,'' in \emph{Proceedings of the IEEE/CVF Conference on Computer Vision and Pattern Recognition}, 2024, pp. 2355--2364.

\bibitem{kazanzides2014open}
P.~Kazanzides, Z.~Chen, A.~Deguet, G.~S. Fischer, R.~H. Taylor, and S.~P. DiMaio, ``An open-source research kit for the da vinci{\textregistered} surgical system,'' in \emph{2014 IEEE international conference on robotics and automation (ICRA)}.\hskip 1em plus 0.5em minus 0.4em\relax IEEE, 2014, pp. 6434--6439.

\bibitem{SARAMIS_Montana_Brown}
N.~Montana-Brown, S.~U. Saeed, A.~Abdulaal, T.~Dowrick, Y.~Kilic, S.~Wilkinson, J.~Gao, M.~Mashar, C.~He, A.~Stavropoulou, E.~Thomson, Z.~M. Baum, S.~Foti, B.~Davidson, Y.~Hu, and M.~Clarkson, ``{SARAMIS}: Simulation assets for robotic assisted and minimally invasive surgery,'' in \emph{Advances in Neural Information Processing Systems}, A.~Oh, T.~Naumann, A.~Globerson, K.~Saenko, M.~Hardt, and S.~Levine, Eds., vol.~36.\hskip 1em plus 0.5em minus 0.4em\relax Curran Associates, Inc., 2023, pp. 26\,121--26\,134.

\bibitem{MONAI_MAISI_2024}
\BIBentryALTinterwordspacing
{NVIDIA MONAI MAISI}, ``{Addressing Medical Imaging Limitations with Synthetic Data Generation},'' 9 2024. [Online]. Available: \url{https://developer.nvidia.com/blog/addressing-medical-imaging-limitations-with-synthetic-data-generation/}
\BIBentrySTDinterwordspacing

\bibitem{DIAZPINTO_MONAI}
A.~Diaz-Pinto, S.~Alle, V.~Nath, Y.~Tang, A.~Ihsani, M.~Asad, F.~P{\'e}rez-Garc{\'\i}a, P.~Mehta, W.~Li, M.~Flores \emph{et~al.}, ``Monai label: A framework for ai-assisted interactive labeling of 3d medical images,'' \emph{Medical Image Analysis}, vol.~95, p. 103207, 2024.

\bibitem{Lorensen_Marching_cubes}
W.~E. Lorensen and H.~E. Cline, ``Marching cubes: A high resolution 3d surface construction algorithm,'' in \emph{Seminal graphics: pioneering efforts that shaped the field}, 1998, pp. 347--353.

\bibitem{qi2017pointnet}
C.~R. Qi, H.~Su, K.~Mo, and L.~J. Guibas, ``{PointNet}: Deep learning on point sets for 3d classification and segmentation,'' in \emph{Proceedings of the IEEE conference on computer vision and pattern recognition}, 2017, pp. 652--660.

\bibitem{ho2020denoisingdiffusionprobabilisticmodels}
J.~Ho, A.~Jain, and P.~Abbeel, ``Denoising diffusion probabilistic models,'' \emph{Advances in neural information processing systems}, vol.~33, pp. 6840--6851, 2020.

\bibitem{zhu2024pointcloudmattersrethinking}
H.~Zhu, Y.~Wang, D.~Huang, W.~Ye, W.~Ouyang, and T.~He, ``Point cloud matters: Rethinking the impact of different observation spaces on robot learning,'' \emph{arXiv preprint arXiv:2402.02500}, 2024.

\bibitem{fls}
\BIBentryALTinterwordspacing
{Fundamentals of Laparoscopic Surgery.} [Online]. Available: \url{https://www.flsprogram.org/}
\BIBentrySTDinterwordspacing

\end{thebibliography}

\end{document}